\documentclass[lettersize,journal]{IEEEtran}
\usepackage{amsmath,amsfonts}
\usepackage{algorithmic}
\usepackage{algorithm}
\usepackage{array}
\usepackage[caption=false,font=normalsize,labelfont=sf,textfont=sf]{subfig}
\usepackage{textcomp}
\usepackage{stfloats}
\usepackage{url}
\usepackage{verbatim}
\usepackage{graphicx}
\usepackage{cite}
\usepackage{xcolor}
\usepackage{multirow}
\usepackage{amsfonts}
\usepackage[colorlinks,bookmarksopen,bookmarksnumbered, citecolor=blue, linkcolor=red, urlcolor=blue]{hyperref}
\usepackage{amsmath}
\usepackage{graphicx}
\usepackage{amssymb}

\begin{document}

\title{Efficient and Robust Knowledge Distillation from A Stronger Teacher Based on Correlation Matching}

\author{Wenqi Niu, Yingchao Wang, Guohui Cai, and Hanpo Hou 
\thanks{This work was supported by the National Key R\&D Program of China under Grant 2021YFB1715700. (\emph{Corresponding author: Yingchao Wang.})} \thanks{Yingchao Wang is with the School of Cyberspace Science and Technology, Beijing Institute of Technology, Beijing, China. (e-mail: yingchaowang@bit.edu.cn)} 
\thanks{Wenqi Niu is with Electronic Information Science and Technology, North Minzu University, Ningxia, Yinchuan, China. (e-mail: niuwenqi\_mail@163.com )}
\thanks{Guohui Cai is with the College of Electronic and Information, Southwest Minzu University, Chengdu, China. (e-mail: guohuicai123@163.com)}
\thanks{Hanpo Hou is with the School of Business, Beijing Technology and Business University, Beijing, China. (e-mail: houhanpo@163.com)}
}
%\thanks{Manuscript received April 19, 2021; revised August 16, 2021.}%


\markboth {}
{Shell \MakeLowercase{\textit{et al.}}: A Sample Article Using IEEEtran.cls for IEEE Journals}


\maketitle
\begin{abstract} Knowledge Distillation (KD) has emerged as a pivotal technique for neural network compression and performance enhancement. Most KD methods aim to transfer dark knowledge from a cumbersome teacher model to a lightweight student model based on Kullback-Leibler (KL) divergence loss. However, the student performance improvements achieved through KD exhibit diminishing marginal returns, where a stronger teacher model does not necessarily lead to a proportionally stronger student model. To address this issue, we empirically find that the KL-based KD method may implicitly change the inter-class relationships learned by the student model, resulting in a more complex and ambiguous decision boundary, which in turn reduces the model's accuracy and generalization ability. Therefore, this study argues that the student model should learn not only the probability values from the teacher’s output but also the relative ranking of classes, and proposes a novel Correlation Matching Knowledge Distillation (CMKD) method that combines the Pearson and Spearman correlation coefficients-based KD loss to achieve more efficient and robust distillation from a stronger teacher model. Moreover, considering that samples vary in difficulty, CMKD dynamically adjusts the weights of the Pearson-based loss and Spearman-based loss. CMKD is simple yet practical, and extensive experiments demonstrate that it can consistently achieve state-of-the-art performance on CIRAR-100 and ImageNet, and adapts well to various teacher architectures, sizes, and other KD methods.
\end{abstract}

\begin{IEEEkeywords}
knowledge distillation, capacity mismatch,  dark knowledge, relaxed distillation, rank relation.
\end{IEEEkeywords}

\section{Introduction}
\label{Section I}
\begin{figure}[!t]
	\centering
	\includegraphics[width=3.5in]{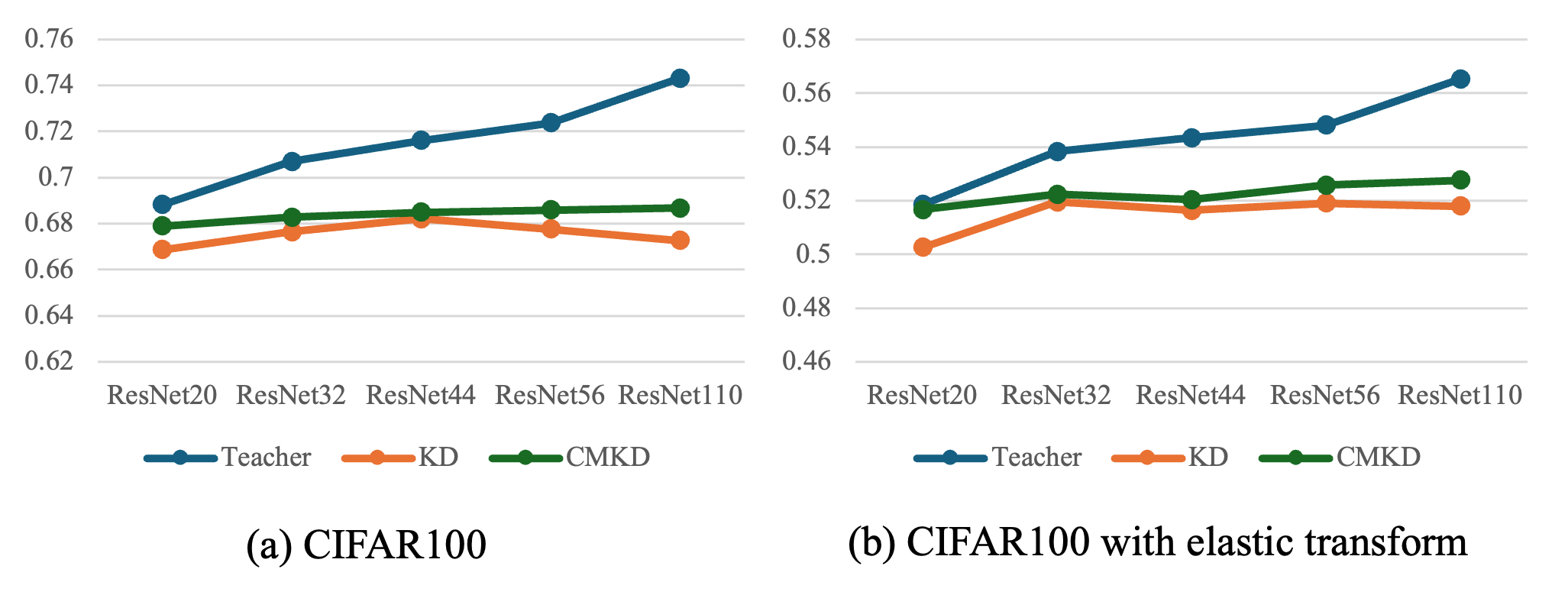}
	\caption{The accuracy of the teacher model and the student model (ResNet14), which are all trained on the clean CIFAR-100 dataset. (a) illustrates the testing Top-1 accuracy on the clean CIFAR-100 dataset, while (b) displays the testing accuracy on the noisy CIFAR-100 dataset with elastic transformations.}
	\label{fig1}
\end{figure}

\IEEEPARstart{I}{n} recent years, Deep Neural Networks (DNNs) have made significant advancements across various fields, particularly in computer vision tasks such as image classification, object detection, and semantic segmentation \cite{yang2022cloud, wang2024towards}. In general, as shown in Figure \ref{fig1}, the accuracy tends to improve as the network size increases, regardless of whether the data is clean or noisy. In other words, larger DNNs (i.e., those with more parameters and deeper layers) tend to exhibit greater accuracy, generalization, and robustness\cite{wang2024end}. However, larger models lead to a corresponding rise in complexity and computational demands, which limits the practical application and deployment of DNNs in resource-constrained environments. 

Knowledge Distillation (KD) \cite{hinton2015distilling} holds the potential to transfer both the accuracy and robustness learned by a larger-scale and higher-capacity teacher model to a smaller and streamlined student model, achieving efficient compression while maintaining commendable performance \cite{gou2021knowledge}. The classical KD process minimizes the Kullback-Leibler (KL) divergence loss between the teacher's output and the student model's output with a fixed temperature \cite{hinton2015distilling}, where the output can be the logits or the softened probabilities. In this way, the student can be guided with more informative signals during training and is thus expected to have a more promising performance than that being trained stand-alone. After years of development, KD has made remarkable progress and has become an effective and well-established paradigm for compressing and enhancing DNNs \cite{song2022spot}.

%figure 1

Intuitively, using a larger and stronger teacher model is expected to distill into a better-performing student model. However, previous studies \cite{cho2019efficacy, park2019relational, mirzadeh2020improved, son2021densely, zhu2021student, wang2022efficient, zhu2022teach, huang2022knowledge, rao2023parameter, liang2024neighbor, yuan2024student,  fanrevisit} have shown that this empiricism does not always hold. The student model distilled from a higher accuracy and larger-scale teacher model may perform worse as shown in Figure \ref{fig1}. This phenomenon has also been observed in different robust distillation methods \cite{yin2024adversarial}. Existing studies attribute the reason behind this phenomenon to the capacity mismatch between the teacher and student models \cite{cho2019efficacy, park2019relational, mirzadeh2020improved, son2021densely, zhu2021student, wang2022efficient, zhu2022teach, huang2022knowledge, rao2023parameter, liang2024neighbor, yuan2024student,  fanrevisit}. To address this issue, some studies \cite{zhu2021student, mirzadeh2020improved,son2021densely} focused on the innovation of KD architecture. For example, TAKD\cite{mirzadeh2020improved} was proposed to reduce the discrepancy between teacher and student by resorting to an additional teaching assistant of moderate model size. On the other hand, some studies aimed to regularize teacher's knowledge to narrow the capacity gap. For example, \cite{wang2022efficient} advocates that an intermediate checkpoint will be more appropriate for distillation. Although these methods provide insights into different aspects, a generic enough solution is preferred to address the difficulty of KD brought by stronger teachers.

Different from the above studies, this study re-examines the reasons why traditional knowledge distillation has poor performance from the perspectives of output-level dark knowledge and inter-class relationships. We empirically find that the KL-based KD method may implicitly change the inter-class relationships learned by the student model, resulting in a more complex and ambiguous decision boundary, which in turn reduces the model's accuracy and generalization ability. Therefore, we demonstrate that enabling the student model to learn the rank relation inherent in the teacher model's output is both sufficient and effective. The rank-based approach \cite{huang2022knowledge, fanrevisit} allows for greater flexibility, improving the student's ability to capture the intrinsic relations in the classes while mitigating the drawbacks associated with KL divergence. Regarding this, we propose a novel Correlation Matching Knowledge Distillation (CMKD) method that combines the Pearson and Spearman correlation coefficients-based KD loss to achieve more efficient and robust distillation from a stronger teacher model. The main contributions of this study can be summarized as follows.

\begin{itemize}
	\item We propose a novel correlation matching KD method (CMKD) that employs a combination of the Pearson and Spearman correlation coefficients to achieve a more flexible alignment between the teacher and student models.

	\item We demonstrated the benefits of relaxed matching and introduced Z-score normalization to approximate a standard normal distribution in the model outputs, thereby satisfying the applicability conditions of the Pearson correlation coefficient.

	\item We assess the difficulty of samples based on the information entropy of the teacher's output and dynamically adjust the weights of the Pearson and Spearman correlation coefficients during the distillation process according to the sample difficulty.
	
\end{itemize}

The rest of this paper is organized as follows. Related studies are reviewed in Section \ref{Section II}. Section \ref{Section III} presents the preliminary knowledge about KD, Pearson and Spearman correlation coefficients. Section \ref{Section IV}  demonstrates the motivation and details of CMKD. Section \ref{Section V} gives the details of the CMKD. Section \ref{Section VI} shows the experiments in different datasets and neural networks and delves into the hyperparameters and ablation experiments. Finally, this paper is concluded in Section \ref{Section VII}.

\section{Related Work}
\label{Section II}
Recently, some studies have been performed to address the poor learning issue of the student model when the student and teacher model sizes significantly differ. Some studies \cite{mirzadeh2020improved, son2021densely, liang2024neighbor, zhu2021student} focused on the innovation of KD architecture. TAKD\cite{mirzadeh2020improved} proposes to reduce the discrepancy between teacher and student by resorting to an additional teaching assistant of moderate model size. DGKD \cite{son2021densely} further improves TAKD by densely gathering all the assistant models to guide the student. NSKD \cite{liang2024neighbor} incorporated teacher assistants into Self-KD by introducing auxiliary classifiers to the shallow layers of the network to reduce the mismatch between the capacities of the student and teacher models. While, SCKD\cite{zhu2021student} investigated the capacity mismatch issue from the perspective of gradient similarity, which dynamically determined when to activate or deactivate the knowledge distillation loss, depending on the relative gradient direction in relation to the student loss. However, these methods require meticulous manual selection of the assistant teacher model or determining the appropriate activate point to achieve an optimal balance in knowledge transfer effectiveness.

On the other hand, some studies \cite{cho2019efficacy, wang2022efficient, zhu2022teach, rao2023parameter, yuan2024student} aimed to regularize teacher's knowledge to narrow the capacity gap. Cho et al. \cite{cho2019efficacy} argued that the KD process can benefit from using an early stopping strategy during training. Similarly, CheckpointKD \cite{wang2022efficient} employed intermediate models from the middle of the training process as teacher models, instead of relying on fully trained models. It further selected an appropriate intermediate teacher model based on mutual information. Zhu et al. \cite{zhu2022teach} demonstrated that the issue of poor learning is directly linked to the presence of undistillable classes. Therefore, they introduced a straightforward "Teach Less, Learn More" framework to identify and exclude these undistillable classes during training. Rao \cite{rao2023parameter} argued that the capacity mismatch issue can be mitigated by ensuring the appropriate smoothness of the soft labels. To achieve this, an adapter module was introduced for the teacher model, where only the adapter is updated to produce soft labels with the desired level of smoothness. SKD \cite{yuan2024student} aims to simplify teacher output into new knowledge representations, which involve softening processing and a learning simplifier. Although these studies have led to improved distillation performance, they have not explored the capacity gap in the context of distillation losses.

Studies \cite{park2019relational, huang2022knowledge, fanrevisit} closely align with our work. RKD \cite{park2019relational} transferred mutual relations of data examples instead, which use distance-wise and angle-wise distillation losses that penalize structural differences in relations. Huang et al. \cite{huang2022knowledge} proposed a Pearson correlation coefficient-based loss to capture the intrinsic inter-class relations from the teacher explicitly. Fan et al.\cite{fanrevisit} observed a positive correlation between the calibration of the teacher model and the KD performance with the original KD methods, and recommended employing measurements insensitive to calibration such as ranking-based loss \cite{huang2022knowledge}. In contrast to the studies mentioned above, we explain the advantages of rank-based KD from the perspective of model decision boundaries and propose using both the Pearson and Spearman correlation coefficients to construct the distillation loss.

\section{Preliminary Knowledge}
\label{Section III}
\subsection{Knowledge Distillation}
In the classic KD method, the transferred knowledge refers to soft labels that are the predictions by the teacher model $ \mathcal T $, and the loss function of the student model $ \mathcal S $ is defined as follows.
\begin{equation}
	\label{Eq1}
	{{\mathcal L_\mathcal S}= \mathcal L_{G}+\mathcal L_{KD}={\mathcal F(\boldsymbol{p,y})}+{\mathcal H(\boldsymbol{p}^{\mathcal T},\boldsymbol{p}^{\mathcal S})}}
\end{equation}

$\mathcal L_{G}=\mathcal F(\boldsymbol{p,y})$ is the cross-entropy loss function between the predicted probability of the student model $\boldsymbol{p}=\left[{{p_1},{p_2},...,{p_c }} \right]$ and the ground truth label $ \boldsymbol{y} \in \left\{{1,\;2,\;...,\;c} \right\}$, where $c$ is the total number of classes, and ${p_i}, i\in \left\{{1,\;2,\;...,\;c}\right\}$ can be obtained by Eq. (\ref{Eq2}). 
\begin{equation}
	\label{Eq2}
	{{p_i} = \frac{{exp\left( {{z^{\mathcal S}_i}} \right)}}{{\mathop \sum \nolimits_{j = 1}^c exp\left( {{z^{\mathcal S}_j}} \right)}}}
\end{equation}
where $z^{\mathcal S}_i$ represents the logit of the $i$-th class from the student model $ \mathcal S $. 

$\mathcal L_{KD}=\mathcal H(\boldsymbol{p}^{\mathcal T}, \boldsymbol{p}^{\mathcal S})$ is the KD loss, which usually is the Kullback-Leibler (KL) divergence loss function between the softened predictions of the student model $\boldsymbol{p}^{\mathcal S}=\left[{{p^{\mathcal S}_1},{p^{\mathcal S}_2},...,{p^{\mathcal S}_c }}\right]$ and the corresponding teacher predictions $\boldsymbol{p}^{\mathcal T}=\left[{{p^{\mathcal T}_1},{p^{\mathcal T}_2},...,{p^{\mathcal T}_c }}\right]$, which is as follows. 
\begin{equation}
	\label{Eq3}
	{{\mathcal H(\boldsymbol{p}^{\mathcal T},\boldsymbol{p}^{\mathcal S})}= \mathop \sum \limits_{i = 1}^c p_i^{\mathcal T} log\left( \frac{p_i^{\mathcal T}}{p_i^{\mathcal S}} \right)}
\end{equation}
\begin{equation}
\label{Eq4}
{{p^{\mathcal T}_i} = \frac{{exp\left( {{z^{\mathcal T}_i}/{T}} \right)}}{{\mathop \sum \nolimits_{j = 1}^c exp\left( {{z^{\mathcal T}_j}/{T}} \right)}}}
\end{equation}
\begin{equation}
\label{Eq5}
{{p^{\mathcal S}_i} = \frac{{exp\left( {{z^{\mathcal S}_i}/{T}} \right)}}{{\mathop \sum \nolimits_{j = 1}^c exp\left( {{z^{\mathcal S}_j}/{T}} \right)}}}
\end{equation}
where $T$ is a temperature coefficient to soften the predicted probability, $z^{\mathcal T}_i$ represents the logit of the $i$-th class from the student model $ \mathcal T $ . 
%figure2 
\begin{figure*}[htbp] 
	\centering
	\includegraphics[width=7.2in]{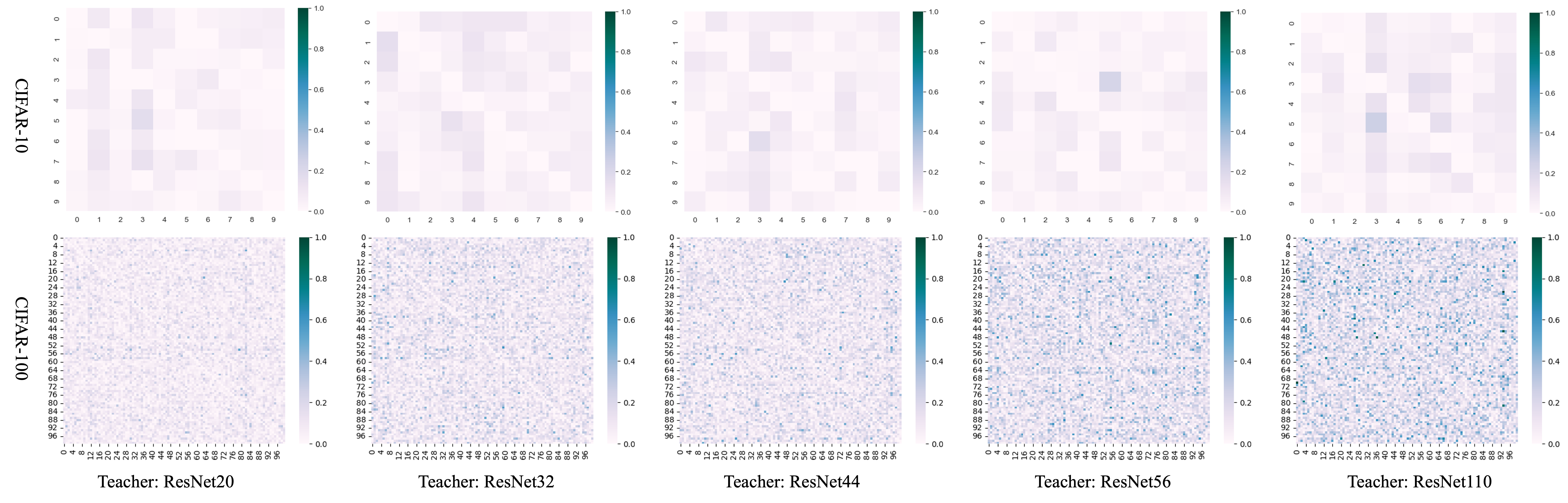}
	\caption{The confusion matrix between the logits of the teacher model and the student model (ResNet14). The first row shows the confusion matrices on the CIFAR-10 dataset, while the second row displays the confusion matrices on the CIFAR-100 dataset.}
	\label{newfig3}
\end{figure*} 

\subsection{Correlation Measures}
\subsubsection{Pearson correlation coefficient}
The Pearson correlation coefficient is a statistical measure that quantifies the linear relationship between two variables $X$ and $Y$. It is defined as the covariance of the two variables divided by the product of their standard deviations. The Pearson correlation coefficient $r$ for two variables $X$ and $Y$ is as follows.
\begin{equation}
	\label{Eq6}
	\resizebox{0.85\hsize}{!}{
		$\begin{aligned}
		{r (X,Y) = \frac{\sum_{i=1}^{n} (X_i - \bar{X})(Y_i - \bar{Y})}{\sqrt{\sum_{i=1}^{n} (X_i - \bar{X})^2} \sqrt{\sum_{i=1}^{n} (Y_i - \bar{Y})^2}}}
		\end{aligned}$}
\end{equation}
where \( X_i \) and \( Y_i \) are the individual sample points, \( \bar{X} \) and \( \bar{Y} \) are the means of the $X$ and $Y$ samples, respectively, and \( n \) is the number of paired observations. 

\subsubsection{Spearman correlation coefficient}
The Spearman correlation coefficient is a non-parametric measure of the strength and direction of the association between two ranked variables. Unlike the Pearson correlation coefficient, Spearman's rank correlation does not assume that the relationship between the variables is linear or that the variables are normally distributed. Instead, it assesses how well the relationship between two variables can be described using a monotonic function. The Spearman correlation coefficient $\rho$ for two variables $X$ and $Y$ is as follows.
\begin{equation}
	\label{Eq7}
\rho(X,Y) = 1 - \frac{6 \sum d_i^2}{n(n^2 - 1)}
\end{equation}
where $d_i=rank(X_i)-rank(Y_i)$ is the difference between the ranks of corresponding values of $X$ and $Y$, $n$ is the number of paired observations. 

%figure 2
%\begin{figure}[!t]
	%\centering
	%\includegraphics[width=3in]{figure/Newfig2.png}
	%\caption{{Different teacher models generate probability distributions for the same sample, which is randomly selected from the CIFAR-100 dataset}}
	%\label{fig2}
%\end{figure} 

%figure 3

\section{Motivation and Theoretical Analysis}
\label{Section IV}
\subsection{Revisit the Capacity Mismatch}
\subsubsection{Capacity mismatch reflected in logit range, and KL-based KD methods cannot reduce the difference between teacher's and student's logits efficiently} The KL-based KD method seeks to align the logit of the student model with that of the teacher model. To assess the similarities and differences between the teacher's logit and the student's logit after KD, we fixed the student model architecture to ResNet14 and employed teacher models of varying capacities (ResNet20, ResNet32, ResNet44, ResNet56, and ResNet110) for KD on the CIFAR-10 and CIFAR-100 datasets, and visualized the confusion matrix between the logits of the teacher model and the student model. As shown in Figure \ref{newfig3}, the color intensity of the confusion matrix reflects the magnitude of the difference between the logits of the teacher and student models, and a darker color signifies a larger discrepancy.

It can be observed that as the size of the teacher model increases, the color of the confusion matrix progressively darkens, reflecting a greater difference as the teacher model's capacity grows. This also illustrates that the KL-based KD method cannot reduce the difference between the logits of the teacher model and the student model. This disparity primarily arises from the capacity gap between the student and teacher models \cite{cho2019efficacy, park2019relational, mirzadeh2020improved, son2021densely, zhu2021student, wang2022efficient, zhu2022teach, huang2022knowledge, rao2023parameter, liang2024neighbor, yuan2024student,  fanrevisit}. Specifically, a more robust teacher model has a stronger representational ability, allowing it to capture complex patterns and relationships in the data more effectively, thereby fitting the training data more accurately and producing sharper output probability distributions. However, due to the smaller capacity of the student model, it is unable to replicate the intermediate features extracted by the teacher model, resulting in the student model's inability to accurately reproduce the teacher's output distribution.

\subsubsection{KL-based exact matching may implicitly change the inter-class relationships learned by the student model} 
\begin{figure*}[!t]
	\centering
	\includegraphics[width=7.2in]{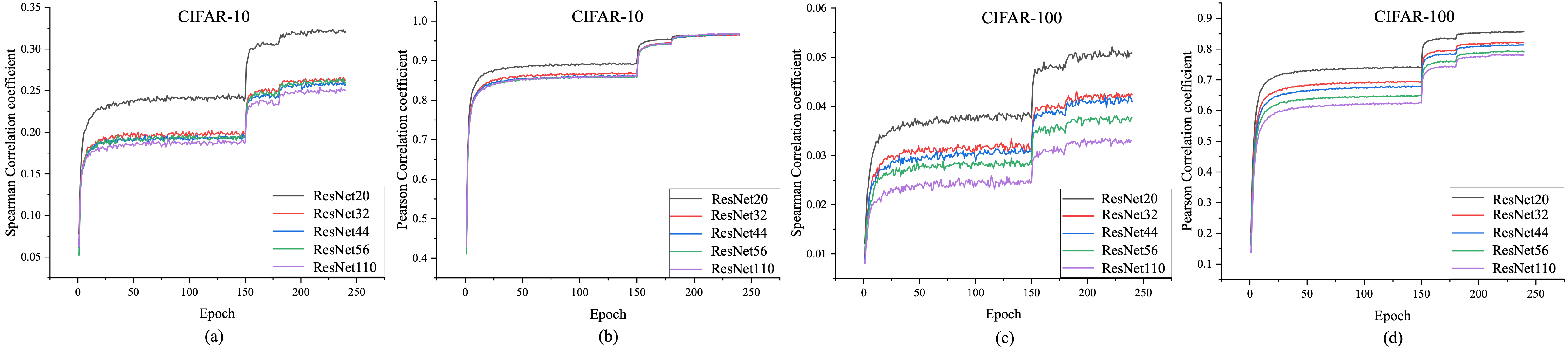}
	\caption{Spearman and Pearson correlation coefficients between the teacher model output and the student model output during knowledge distillation. The training dataset for (a) and (b) is CIFAR-10, while the training dataset for (c) and (d) is CIFAR-100.}
	\label{Newfig5}
\end{figure*}

The rank relationship of a model's output is determined by comparing the output values (logits) for each class. Previous study \cite{li2024exploring} has demonstrated that while more powerful teacher models tend to produce probability vectors with smaller distinctions between non-target classes, teachers of varying capacities generally maintain consistent perceptions of relative class affinities. However, KL-based knowledge distillation is ineffective at capturing the rank relationships of the teacher model. 
As shown in Figure \ref{Newfig5}, with the increase in training iterations, the Spearman correlation coefficient between the outputs of the student model and the teacher model gradually increases, but remains in a state of low correlation. This is particularly evident on CIFAR-100, where the rank relationship between the two models shows almost no correlation. Furthermore, the larger the teacher model, the lower the correlation between the rank of the outputs from the teacher and student models.

%figure 3

\begin{figure}[!t]
	\centering
	\includegraphics[width=3.5in]{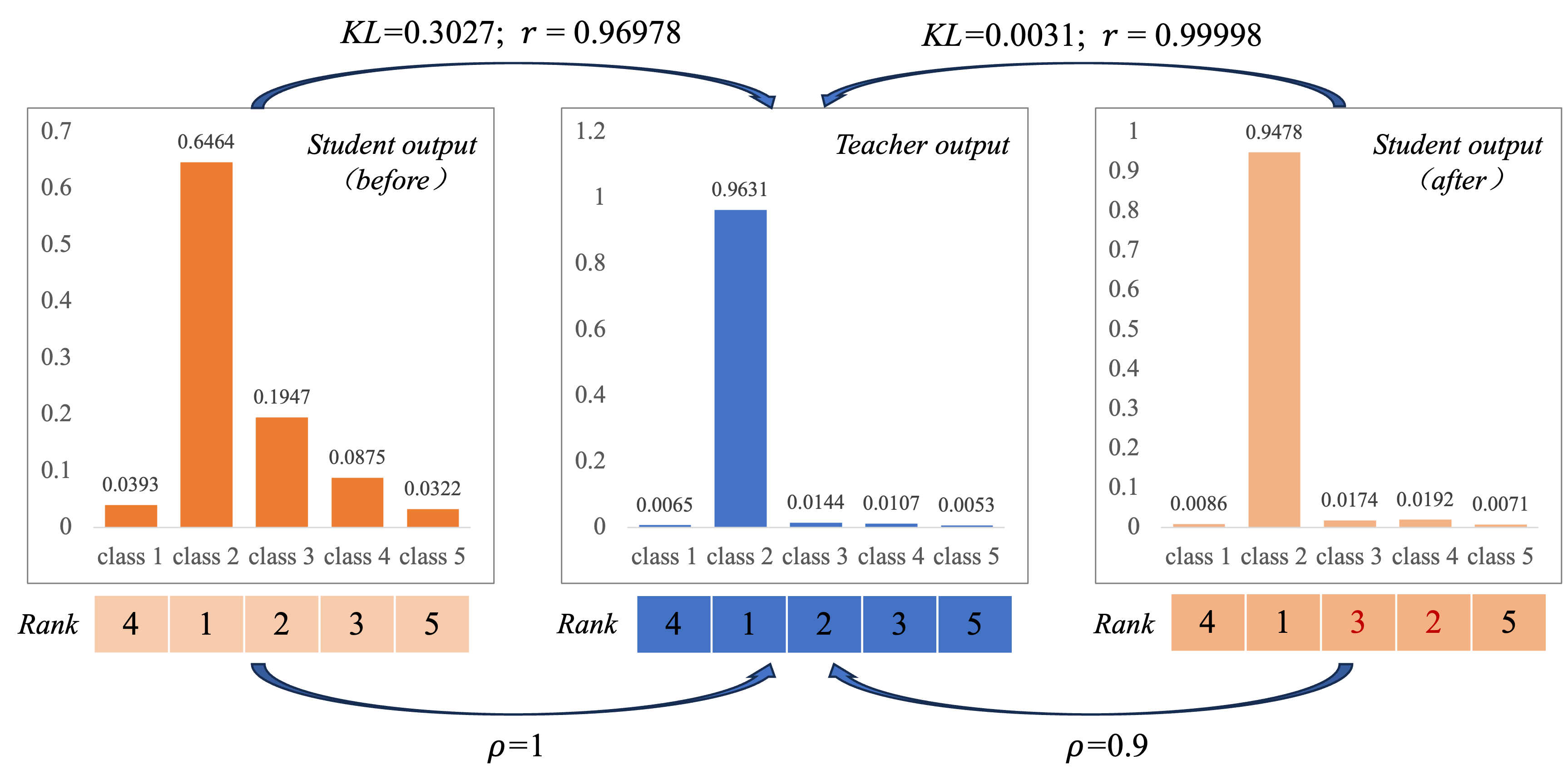}
	\caption{An example of implicitly altering the rank relationship of the student model's output through the KL-based KD method, where $r$ is the Pearson correlation coefficient and $\rho$ is the Spearman correlation coefficient.}
	\label{fig4}
\end{figure}

To illustrate the above phenomenon, we derive the KL loss function $\mathcal{L}_{KD}$ of the student model with respect to the logits $z_k^s$ as Eq.(\ref{eq8}). The specific derivation process can be found in the Appendix.
\begin{equation}
    \label{eq8}
    \frac{\partial \mathcal{L}_{KD}}{\partial z_k^\mathcal S}  = \frac{1}{T}(p_{k}^{\mathcal S}-p_{k}^{\mathcal T})
\end{equation}
It reveals that, in the KD process, for any input sample belonging to class $k$, the direction and magnitude of the gradient update resulting from matching via KL divergence are determined by the discrepancy between the student model's output $p_{k}^{\mathcal S}$ and the teacher model's output $p_{k}^{\mathcal T}$. However, as shown in Figure \ref{newfig3}, the varying color intensities across different categories indicate that the differences between the teacher and student model outputs are not consistent across all classes, resulting in the KL-based method prioritizing fitting the classes with large differences in the logit value, which may alter the relative rank relationships among classes with smaller probability differences. For example, as shown in Figure \ref{fig4}, during the training process of the student model, to minimize the KL divergence loss, the probability of the target class 2 will increase preferentially, while the probabilities of the other non-target classes will decrease. However, due to the varying differences between the output probabilities of each class in the student and teacher models, the probabilities of the non-target classes in the student model decrease at different rates, which affects the rank order of the non-target classes (e.g., in the teacher model, class 4 has a higher rank than class 5, but the student model may learn an order where class 4 has a lower rank than class 3).

%figure 4

\subsubsection{The changes in relative ranks among non-target classes will compel the student model's decision boundaries to shift and become more complex and ambiguous} The rank relationship of the model’s output is closely related to its decision boundary. A higher rank indicates greater confidence in a particular class, and the model is more likely to assign the corresponding sample to that class along the decision boundary. For example, as shown in Figure \ref{fig4}, if the relative rank order between class 3 and class 4 is altered, the student model may perceive class 2 samples to be more similar to class 4 samples than to class 3 samples. This could potentially shift the decision boundaries between class 2 and class 3, as well as between class 2 and class 4, leading to blurred and more complex decision boundaries. As a result, the student model may find it more difficult to learn robust features, potentially reducing its generalization and robustness ability. 

\begin{figure*}[!t]
	\centering
	\includegraphics[width=7in]{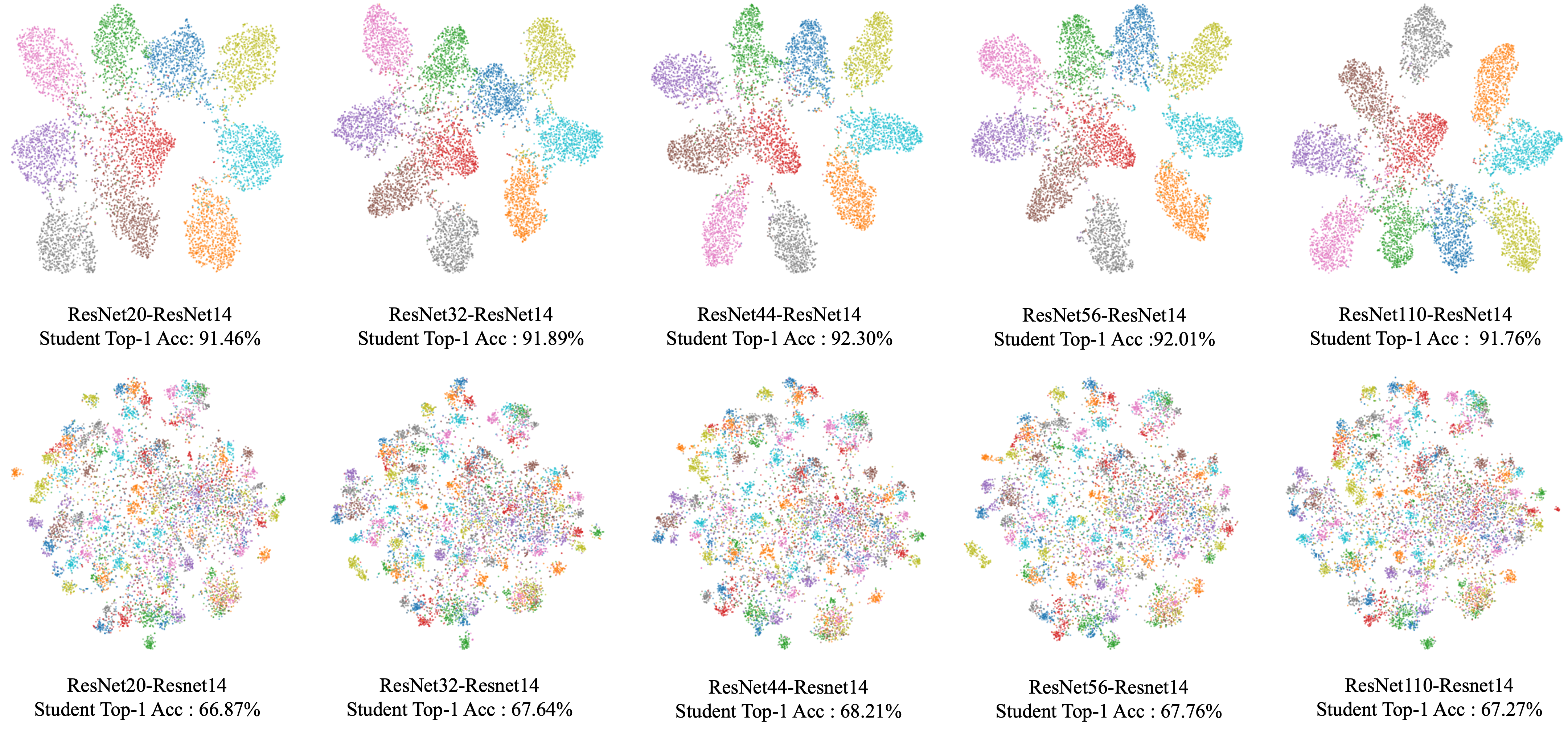}
	\caption{T-SNE dimensionality reduction visualization for the same student under different teachers. The first row shows the results on the CIFAR-10 dataset, and the second row shows the results on the CIFAR-100 dataset.}
	\label{newfig4}
\end{figure*}

To illustrate the presence of boundary blurring, we conducted experiments on the CIFAR-10 and CIFAR-100 dataset, using different teacher models (ResNet20, ResNet32, ResNet44, ResNet56, and ResNet110) to guide the training of a student model (ResNet14), and employed t-SNE for visualization, where different colors represent different categories in the classification. The more concentrated the clusters of the same color and the more dispersed the clusters of different colors, the stronger the model's discriminative capability and the clearer its decision boundaries. As shown in Figure \ref{newfig4}, as the capacity of the teacher model increases, the clustering boundary between the red class and other classes initially becomes clearer but later becomes blurred. The clearest boundary is observed when the teacher model is ResNet44. This indicates that when the capacity of the teacher model is too large, the decision boundaries of the student model tend to become complex and blurred, further supporting the analysis presented above.

%figure 6

Therefore, this paper proposes that the student model should learn not only the probability distribution values from the teacher model but also the relative rank relationships among classes in the teacher's output.

\subsection{Explore Relaxed Matching based on Rank Relations} 
\subsubsection{Linear correlation}
The Pearson correlation coefficient \cite{pearson1896vii} measures the linear correlation between two variables, thus reflecting their rank relationships. Huang et al.\cite{huang2022knowledge} have introduced Pearson correlation as a substitute for KL divergence, encouraging the student model’s output to be as positively correlated as possible with that of the teacher model. However, as shown in Figure \ref{Newfig5}, the Pearson coefficient between the teacher model and the student model is relatively high, indicating a strong linear correlation. In reality, the rank correlation between the two models remains low. This suggests that while the Pearson coefficient aids the student model in approximating the numerical values of the teacher model’s outputs, it does not effectively help the student model in learning the rank relationship of the teacher model’s outputs. 

Specifically, on the one hand, Pearson correlation is mainly suited to capturing linear relationships. Given the disparity in capacities between the teacher and student models, the student model cannot precisely replicate the teacher model’s output distribution. This results in a relationship between the teacher and student model outputs that is not always linear, and Pearson correlation may fail to accurately capture such non-linear relationships, leading to suboptimal knowledge distillation performance. On the other hand, the Pearson correlation is highly sensitive to outliers. When the teacher model’s output is overly sharp (especially for simple samples), excessively high probability values may significantly affect the coefficient's calculation, compromising the stability and effectiveness of the knowledge distillation process. For example, as shown in Figure \ref{fig4}, the Pearson correlation coefficient $r$ between the outputs of the student and teacher models changes from 0.96978 to 0.99998, indicating only a small change, which fails to effectively capture the shifts in the rank relationships among the non-target classes.

\subsubsection{Non-linear correlation}
Spearman's rank correlation coefficient, another commonly used metric for assessing rank relationships between two variables, evaluates their monotonic relationship by comparing their ranks without requiring the relationship to be linear. It is also less sensitive to outliers, making it a more relaxed measure of correlation. However, Spearman's coefficient only considers the rank order of outputs, ignoring the actual magnitudes of the values, and therefore does not provide effective guidance in terms of feature extraction.

Therefore, this study proposes to jointly apply Pearson and Spearman coefficients as distillation loss functions, dynamically adjusting their weights based on the difficulty of the samples. Specifically, for simple samples, a highly complex teacher model may overfit the training data, resulting in sharper output probability distributions that capture the details and noise in the training set. In contrast, a simpler student model may be better suited to handling these simple samples, as it is more adept at capturing the fundamental patterns and structure in the data without being distracted by noise. Therefore, for simple samples, we propose assigning a higher weight to the loss based on Spearman's coefficient, ensuring that the model learns the rank relationships from the teacher model while preserving the student model’s own insights regarding probability values. For more difficult samples, we propose assigning a higher weight to the distillation loss based on Pearson's coefficient, so that the student model learns not only the rank knowledge from the teacher model but also pays closer attention to the value-based knowledge, enabling it to capture the complex patterns and relationships in the data.

\section{Methodology}
\label{Section V}
\subsection{Z-score Normalization}
The Pearson correlation coefficient assumes that data follows a normal distribution. However, as shown in Figure \ref{figNew7}, the logit of both the teacher and student models resemble a normal distribution, but not completely normal. To address this, we applied Z-score normalization to the logit of the teacher and student models, ensuring that the logits conform to a standard normal distribution without altering the relationships between their outputs. Z-score normalization primarily adjusts the scale of the data without changing the relative positions or rank order of the data. In other words, it does not alter the nonlinear relationships between the data and does not affect the use of Spearman's rank correlation coefficient. Moreover, Z-score normalization can reduce the magnitude and variance differences between the logits of the teacher and student models, thereby mitigating the negative impact of logit value discrepancies on the distillation process.

The calculation formula of the Z-score normalization is as follows.
\begin{equation}
	\label{Eq:9}
    {\hat{z}_{i}=\frac{z_{i}-\mu}{\sigma}}
\end{equation}
where $\mu$ and $\sigma$ are the mean and variance of the model logits output, respectively, and the calculation formula is as follows.
\begin{equation}
	\label{Eq:10}
    {\mu=\frac{1}{c}\sum_{i=1}^{c}z_{i}}
\end{equation}

\begin{equation}
	\label{Eq:11}
    {\sigma=\sqrt{\frac{1}{c-1}\sum_{i=1}^{c}(z_{i}-\mu)^{2}}}
\end{equation}
where $c$ represents the number of categories, and $z_{i}$ represents the logits output value of the model for the $i$-th category.

\begin{figure}
	\centering
	\includegraphics[width=0.8\linewidth]{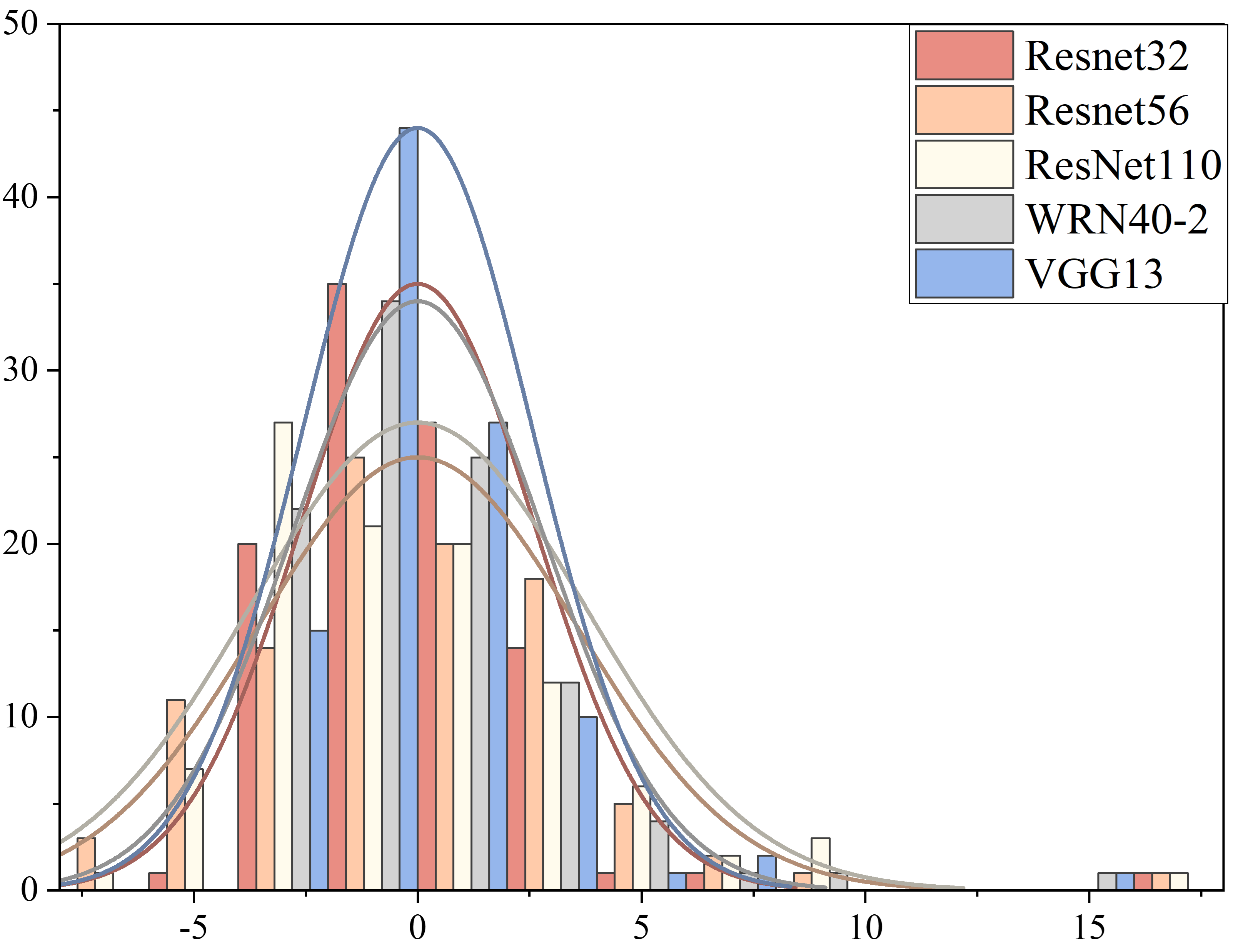}
	\caption{The distributions of logits from teacher models with different architectures on the CIFAR-100 dataset are similar to a normal distribution, but they do not fully conform to a standard normal distribution.}
	\label{fig-3-3(a)}
\end{figure}

\subsection{Correlation Matching}
\subsubsection{Linear Correlation Matching with Pearson Correlation Distillation}
The Pearson correlation distillation loss aims to ensure that the outputs of the student model are as positively correlated with the teacher model's outputs as possible. The loss function is defined as follows.
\begin{equation}
	\label{Eq:12}
    {\mathcal L_{Person}=1-r(\boldsymbol{p}^{\mathcal T}, \boldsymbol{p}^{\mathcal S})}
\end{equation}
where \(r(\boldsymbol{p}^{\mathcal T}, \boldsymbol{p}^{\mathcal S})\) represents the Pearson correlation coefficient between the teacher model's output $\boldsymbol{p}^{\mathcal T} $ and the student model's output $\boldsymbol{p}^{\mathcal S}$.

\subsubsection{Non-linear Correlation Matching with Spearman Correlation Distillation}
The Spearman correlation distillation loss aims to ensure that the output rank relationship of the student model closely aligns with that of the teacher model. The loss function is defined as follows.
\begin{equation}
	\label{Eq:13}
\mathcal L_{Spearman} = 1 - \rho (\boldsymbol{p}^{\mathcal T}, \boldsymbol{p}^{\mathcal S})
\end{equation}
where \(\rho (\boldsymbol{p}^{\mathcal T}, \boldsymbol{p}^{\mathcal S})
\) represents the Spearman correlation between the teacher model's output $P^{\mathcal T} $ and the student model's output $P^{\mathcal S}$. However, the Spearman correlation coefficient requires ranking operations, which are mathematically neither directly differentiable nor tractable for gradient-based optimization. Fortunately, Blondel et al. \cite{blondel2020fast} have proposed a fast and differentiable sorting and ranking method, which is adopted in this study.

\subsection{Dynamic Relaxation Matching Based on Sample Difficulty}

Information entropy is a fundamental measure of uncertainty and information content in probability distributions. Therefore, in this study, the entropy of the teacher model's output is used to measure the difficulty of the samples or the sharpness of the teacher model's output. Specifically, when the teacher model is highly confident in certain simple samples, resulting in a sharper output distribution, and the corresponding entropy is lower. Conversely, when the output distribution is flatter, the entropy is higher, indicating that the sample is more challenging. The formula for calculating the entropy of the model's output is as follows.
\begin{equation}
	\label{Eq:14}
    { H(z)=-\sum_{c=1}^{c}p(z_{c})\log p(z_{c}) }
\end{equation}
where  \(C\) is the total number of classes in the model's output, \(z_{c}\) represents the logits value of the \(c\)-th class, and \(p(z_{c})\) is the probability of the \(c\)-th class in the model's output.

To evaluate whether the output of the teacher model is sharp, we use the average entropy of the teacher model's outputs within a batch as the threshold criterion. The formula for calculating the average entropy is as follows.
\begin{equation}
	\label{Eq:15}
    {\overline{H}(z)=\frac{1}{N}\sum_{i=1}^{N}H_{i}(z) }
\end{equation}

Where \(N\) represents the total number of samples in the batch. If the output entropy of a sample within a batch exceeds the average entropy, we define the teacher model’s output for that sample as excessively sharp. In this case, the distillation loss function primarily calculates the Spearman correlation coefficient, with the Pearson correlation coefficient serving as a supplementary measure. Conversely, if the output entropy of a sample is below the average, we define the teacher model’s output as relatively flat. In this scenario, the distillation loss function primarily focuses on calculating the Pearson correlation coefficient, with the Spearman correlation coefficient as a supplementary measure. Therefore, the loss function during training is defined as follows.

\begin{equation}
    \label{Eq:16}
    L=\begin{cases}
        \alpha L_{CE} + \beta L_{Pearson} + \gamma L_{Spearman}, & \text{if } H_{i} \geq \overline{H} \\
        \alpha L_{CE} + \gamma L_{Pearson} + \beta L_{Spearman}, & \text{if } H_{i} < \overline{H}
    \end{cases}
\end{equation}
where \(\alpha\), \(\beta\), and \(\gamma\) are three hyperparameters used to balance the weights of the target loss and distillation loss in the total loss function. \(H_{i}\) represents the entropy of the \(i\)-th sample in a given batch during training, while \(\overline{H}\) denotes the average entropy of all samples in the batch.

\section{Experiment}
\label{Section VI}

\subsection{Experiment Setting}
\subsubsection{Datasets} To provide a detailed comparison, we conducted experiments on two popular datasets, CIFAR-100 \cite{2009Learning} and ImageNet \cite{deng2009imagenet}. CIFAR-100 \cite{2009Learning} is an image classification dataset consisting of 100 classes, with each class containing 600 images at a resolution of 32$\times$32.  Due to its diverse classes and relatively small image size, CIFAR-100 \cite{2009Learning} is a popular choice for studying image classification and knowledge distillation (KD) methods. ImageNet \cite{deng2009imagenet}, being one of the largest image classification datasets, contains 1,000 classes and over 1.2 million images.  The wide range of categories and large-scale images in ImageNet provides an excellent test environment for evaluating a model's generalization ability and robustness.

\subsubsection{Network Architectures} We employed various popular network architectures as teacher and student models, including VGG\cite{simonyan2014very}, ResNet\cite{he2016deep}, WideResNet\cite{zagoruyko2016wide} series, and lightweight networks such as the MobileNet\cite{sandler2018mobilenetv2} and ShuffleNet \cite{zhang2018shufflenet} series, to assess the performance of our method across different architectures. Additionally, we considered heterogeneous teacher-student model configurations with different network architectures.

\subsubsection{Performance Comparison} In terms of performance comparison, we not only benchmarked our method against the standard KL-based KD \cite{hinton2015distilling} but also compared it with other prior studies. These included logit-based distillation methods such as TAKD \cite{mirzadeh2020improved}, DKD \cite{zhao2022decoupled}, DIST \cite{huang2022knowledge}, and NKD \cite{yang2023knowledge}, as well as feature-based KD methods including FitNet\cite{romero2014fitnets}, AT \cite{zagoruyko2016paying}, RKD\cite{park2019relational}, OFD\cite{heo2019comprehensive}, CRD \cite{tian2019contrastive}, and ReviewKD \cite{chen2021distilling}.

\subsubsection{Implementation Details} We strictly adhered to the experimental settings from prior studies to ensure consistency in hyperparameters such as learning rate, batch size, and optimizer. For the CIFAR-100 dataset, we set the batch size to 64 and the weight decay factor to $5\times10^{-4}$. All models, except for the MobileNet and ShuffleNet series, were initialized with a learning rate of 0.05; for the MobileNet and ShuffleNet series, the initial learning rate was set to 0.01. The training process lasted for 240 epochs, with the learning rate decaying by a factor of 0.1 at the 150th, 180th, and 210th epochs. For the ImageNet dataset, we used a batch size of 512 and a weight decay factor of $1\times10^{-4}$. The training period was 100 epochs, with an initial learning rate of 0.2, which decayed by a factor of 10 at the 30th, 60th, and 90th epochs. 

Across all datasets, we used SGD as the optimizer, with a momentum parameter of 0.9. For the CIFAR-100 dataset, we set $\alpha=1$, $\beta=4$, and $\gamma=1$ with a temperature \(T=4\). For the ImageNet dataset, we similarly set  $\alpha=1$, $\beta=4$, and $\gamma=1$, but with a temperature \(T=1\). All experiments were conducted on an NVIDIA 3080 GPU. The CIFAR-100 dataset was trained on a single GPU, while ImageNet was trained on four GPUs. 

\subsubsection{Evaluation Metrics} 
We used Top-1 and Top-5 accuracy for classification tasks as the primary evaluation metrics, and the final reported results are based on the average of three experimental runs. In CIFAR-100, we relied on Top-1 accuracy as the main metric, while for ImageNet, both Top-1 and Top-5 accuracy are employed. Additionally, we recorded training time to compare the computational cost of our method.

\subsection{Experimental Results}
\begin{table*}
    \centering
    \caption{The Top-1 accuracy of different knowledge distillation methods on the CIFAR-100 validation set is compared. The teacher and student models share the same architecture but have different configurations. In this comparison, $\triangle_{1}$ represents the performance improvement of NormKD relative to classical KD, while $\triangle_{2}$ indicates the performance improvement of DKD + our method compared to DKD.}

    \resizebox{0.9\textwidth}{!}{
    \begin{tabular}{cccccccc} 
    \hline
    \multirow{4}{*}{Distillation      methods} & \multirow{2}{*}{Teacher}            & ResNet32$\times$4 & ResNet56 & ResNet110 & WRN-40-2 & WRN-40-2 & VGG13  \\
    &  & 79.42 & 72.37 & 74.31     & 75.61    & 75.61    & 74.64  \\
    & \multirow{2}{*}{Student}            & ResNet8$\times$4  & ResNet20 & ResNet32  & WRN-40-1 & WRN-16-2 & VGG8   \\
    & & 72.50      & 69.06    & 71.14     & 71.98    & 73.26    & 70.36  \\ 
    \hline
    \multirow{7}{*}{Feature-based methods}     & FitNet[2014]                        & 73.50      & 69.21    & 71.06     & 72.24    & 73.58    & 71.02  \\
        & AT[2016]                            & 73.44      & 70.55    & 72.31 & 72.77    & 74.08    & 71.43  \\
        & VID[2019]                           & 73.09      & 70.38    & 72.61     & 73.30    & 74.11    & 71.23  \\
        & RKD[2019]                           & 71.90      & 69.61    & 71.82     & 72.22    & 73.35    & 71.48  \\
        & OFD[2019]                           & 74.95      & 70.98    & 73.23     & 74.33    & 75.24    & 73.95  \\
        & CRD[2020]                           & 75.51      & 71.16    & 73.48     & 74.14    & 75.48    & 73.94  \\
        & ReviewKD[2021]                      & 75.63      & 71.89    & 73.89     & 75.09    & 76.12    & 74.84  \\ 
    \hline
    \multirow{9}{*}{Logit-based methods}      & KD[2015]                            & 73.33      & 70.66    & 73.08     & 73.54    & 74.98    & 72.98  \\
    & TAKD[2020]                              & 73.81      & 70.83    & 73.37     & 73.78    & 75.12    & 73.23  \\
    & DKD[2022]                               & 76.32      & 71.97    & 74.11     & 74.81    & 76.24    & 74.68  \\
   & DIST[2022]                               & 76.16      & 71.55    & 73.55     & 74.42    & 75.29    & 73.74  \\
    & NKD[2023]                               & 76.35      & 71.62    & 73.79     & 75.23    & 76.37    & 74.86  \\ 
    \cline{2-8}
    & Ours (CMKD)                            & 76.89      & 71.83    & 74.03     & 74.67    & 76.60    & 74.32  \\
    & {$+\triangle_{1}$}                     & +3.56      & +1.17    & +0.95     & +1.13    & +1.62    & +1.34  \\
    & DKD+Ours (CMKD)                        & 77.13      & 72.26    & 74.31     & 75.02    & 76.82    & 74.51  \\
    & {$+\triangle_{2}$}                     & +0.81      & +0.29    & +0.20     & +0.21    & +0.58    & -0.17  \\
    \hline
    \end{tabular}}

\label{tab:4-1-1}
\end{table*}

\begin{table*}
    \centering
    \caption{The Top-1 accuracy of different knowledge distillation methods on the CIFAR-100 validation set is compared, where the teacher and student models have different architectures and configurations. In this comparison, $\triangle_{1}$ represents the performance improvement of NormKD relative to classical KD, while $\triangle_{2}$ indicates the performance improvement of DKD + our method compared to DKD.}
    \resizebox{\textwidth}{!}{%
    \begin{tabular}{cccccccc} 
    \hline
    \multirow{4}{*}{distillation      methods} & \multirow{2}{*}{Teacher} & ResNet32$\times$4   & WRN-40-2     & ResNet32$\times$4   & ResNet50     & VGG13        & WRN-40-2   \\
                                               &                          & 79.42        & 75.61        & 79.42        & 79.34        & 74.64        & 75.61      \\
                                               & \multirow{2}{*}{Student} & ShufleNet-V1 & ShufleNet-V1 & ShufleNet-V2 & MobileNet-V2 & MobileNet-V2 & ResNet8x4  \\
                                               &                          & 70.50        & 70.50        & 71.82        & 64.60        & 64.60        & 72.50      \\ 
    \hline
    \multirow{7}{*}{Feature-based methods}     & FitNet[2014]             & 73.59        & 73.73        & 73.54        & 63.16        & 64.16        & 74.61      \\
                                               & AT[2016]               & 71.73        & 73.32        & 72.73        & 58.58        & 59.40        & 74.11      \\
                                               & VID[2019]                & 73.38        & 73.61        & 73.57        & 65.79        & 65.56        & 74.65      \\
                                               & RKD[2019]                & 72.28        & 72.21        & 73.21        & 64.43        & 64.52        & 75.26      \\
                                               & OFD[2019]                & 75.98        & 75.85        & 76.82        & 69.04        & 69.48        & 74.36      \\
                                               & CRD[2020]                & 75.11        & 76.05        & 75.65        & 69.11        & 69.73        & 75.24      \\
                                               & ReviewKD[2021]           & 77.45        & 77.14        & 77.78        & 69.89        & 70.37        & 74.34      \\ 
    \hline
    \multirow{9}{*}{Logits-based methods}      & KD[2015]                 & 74.07        & 74.83        & 74.45        & 67.35        & 67.37        & 73.79      \\
                                               & TAKD[2020]               & 74.53        & 75.34        & 72.12        & 68.02        & 67.91        & 74.03      \\
                                               & DKD[2022]                & 76.45        & 76.70        & 77.07        & 70.35        & 69.71        & 75.56      \\
                                               & DIST[2022]               & 75.23        & 75.23        & 77.35        & 69.14        & 68.48        & 75.67      \\
                                               & NKD[2023]                & 75.31        & 75.96        & 76.26        & 69.39        & 68.72        & 76.01      \\ 
    \cline{2-8}
                                               & Ours (CMKD)               & 75.71        & 76.72        & 76.48        & 69.59        & 69.23        & 76.96      \\
                                               & $+\triangle_{1}$          & +1.64        & +1.89        & +2.03        & +2.24        & +1.86        & +3.17      \\
                                               & DKD+Ours (CMKD)           & 76.98        & 77.01        & 77.69        & 70.37        & 69.60        & 77.16      \\
                                               & $+\triangle_{2}$          & +0.53        & +0.31        & +0.62        & +0.02        & -0.11        & +1.60      \\
    \hline
    \end{tabular}
    }
   
    \label{tab:4-1-2}
    \end{table*}

\subsubsection{Results on CIFAR-100} We reported the results in Tables \ref{tab:4-1-1} and \ref{tab:4-1-2}. Table \ref{tab:4-1-1} focuses on teacher/student models with the same architecture, while Table \ref{tab:4-1-2} explores combinations with different architectures. As shown in Table \ref{tab:4-1-1}, although feature-based distillation methods generally outperform logits-based methods, our logits-based distillation approach CMKD demonstrated significant performance improvements. In some cases, its performance was on par with or surpassed feature-based methods. Notably, in the combinations of ResNet32$\times$4 with ResNet8$\times$4 and WRN-40-2 with WRN-16-2, our method improved Top-1 accuracy by 3.56\% and 1.62\%, respectively, compared to traditional KD. On the other hand, as shown in Table \ref{tab:4-1-2}, CMKD also achieved significant results in heterogeneous networks. For example, in the combinations of WRN-40-2 with ResNet8$\times$4 and ResNet50 with MobileNet-V2, our method improved Top-1 accuracy by 3.17\% and 2.24\%, respectively, over traditional KD. 

In addition, CMKD can integrate smoothly with other logit-based methods while maintaining simplicity. The results at the bottom of Tables \ref{tab:4-1-1} and \ref{tab:4-1-2} demonstrate that when combined with the DKD method, the performance of DKD improved significantly. In the combinations of ResNet32$\times$4 with ShuffleNet-V1 and ResNet32$\times$4 with ShuffleNet-V2, our method CMKD combined with DKD, further improved Top-1 accuracy by 1.27\% and 1.21\%, respectively. Meanwhile, the results in models with different architectures came closer to the feature-based ReviewKD, and in models with the same architecture, CMKD combined with DKD performed even better. 

\subsubsection{Results on ImageNet} We used ResNet34 as the teacher model and ResNet18 as the student model to form combinations with the same architecture. Similarly, ResNet50 was used as the teacher model and MobileNetV1 as the student model to form combinations with different architectures. As shown in Tables \ref{tab:4-2-1} and \ref{tab:4-2-2}, our method CMKD achieved significant improvements in both Top-1 and Top-5 accuracies. Specifically, for the same architecture combination of ResNet34/ResNet18, compared to traditional KD, CMKD improved Top-1 accuracy by 1.36\% and Top-5 accuracy by 0.84\%. Compared to Review-KD and DKD, CMKD increased Top-1 accuracy by 0.41\% and 0.32\%, and Top-5 accuracy by 0.42\% and 0.52\%, respectively. On the other hand, for the different teacher-student architecture combinations (ResNet50/MobileNetV1), CMKD also performed well, showing improvements of 3.84\% and 0.37\% in Top-1 accuracy compared to KD and DKD, respectively. 

Furthermore, combining our method with DKD could further enhance model performance, resulting in additional gains of 0.19\% and 0.69\% in Top-1 accuracy, and 0.21\% and 0.33\% in Top-5 accuracy, respectively, for the ResNet34/ResNet18 and ResNet50/MobileNetV1 combinations.

\begin{table*}
    \centering
    \caption{Performance comparison of different KD methods on the ImageNet with the same teacher-student architecture (ResNet34-ResNet18) in terms of Top-1 and Top-5 accuracy. $+\triangle_{1}$ represents the performance improvement of CMKD over classical KD, and $+\triangle_{2}$ represents the performance improvement of DKD + CMKD over DKD.}
    \resizebox{\textwidth}{!}{%
    \begin{tabular}{c|ccc|cccc|cccccc} 
    \hline
    \multicolumn{4}{c|}{Distillation Methods}                                 & \multicolumn{4}{c|}{Feature-based methods}                       & \multicolumn{6}{c}{Logit-based methods}                                                   \\                
    \hline
    Teacher - Student                         & Accuracy & Teacher & Student & AT & OFD & CRD & Review KD & KD & DKD & CMKD  & $+\triangle_{1}$ & DKD+CMKD & $+\triangle_{2}$  \\
    \multirow{2}{*}{ResNet34 - ResNet18}      & Top-1    & 73.31   & 69.75   & 70.69    & 70.81     & 71.17     & 71.61           & 71.03    & 71.70     & 72.02 & +0.99            & 72.21    & +0.51 \\ 
    	& Top-5    & 91.42   & 89.07   & 90.01    & 89.98     & 90.13     & 90.51 & 90.05  & 90.41 & 90.72 & +0.67  & 90.93 & +0.52 \\
    \hline
    \end{tabular}
    }
\label{tab:4-2-1}
\end{table*}

\begin{table*}
    \centering
    \caption{Comparison of different KD methods on the ImageNet with the different teacher-student architectures (ResNet50-MobileNetV1) in terms of Top-1 and Top-5 accuracy. $+\triangle_{1}$ represents the performance improvement of CMKD over classical KD, and $+\triangle_{2}$ represents the performance improvement of DKD+CMKD over DKD.}
    \resizebox{\textwidth}{!}{%
    \begin{tabular}{c|ccc|cccc|cccccc} 
    \hline
    \multicolumn{4}{c|}{distillation methods}                              & \multicolumn{4}{c|}{Features}                       & \multicolumn{6}{c}{Logitrs}                                                                                      \\ 
    \hline
    Teacher - Student & Accuracy & teacher & student & AT & OFD & CRD & Review KD & KD & DKD & CMKD  & $+\triangle_{1}$ & DKD+CMKD & $+\triangle_{2}$  \\
    \multirow{2}{*}{ResNet50-MobileNet-V1} & Top-1    & 76.16   & 68.87   & 69.56    & 71.25     & 71.37     & 72.56           & 70.50    & 72.05     & 72.42 & +1.92            & 73.11    & +1.06  \\
       & Top-5    & 92.86   & 88.76   & 89.33    & 90.34     & 90.41     & 91.00           & 89.80    & 91.05     & 90.83 & +1.03                                                & 91.16    & +0.11  \\
    \hline
    \end{tabular}
    }
    \label{tab:4-2-2}
    \end{table*}

\subsubsection{Training Time Comparison} To demonstrate the simplicity and effectiveness of our method, we evaluated the training times of several state-of-the-art distillation techniques to assess the training efficiency of our approach. As shown in Figure \ref{fig:4-2-3}, the training time of our method is comparable to that of traditional KD and significantly less than several feature-based distillation methods. This is because our method only modifies the loss function and does not introduce additional complex structures or computational rules. These results prove that our method is simple and has high training efficiency.

\begin{figure}[htbp] % Adjusts the position of the figure: h (here), t (top), b (bottom), p (on its own page)
	\centering
	\includegraphics[width=\linewidth]{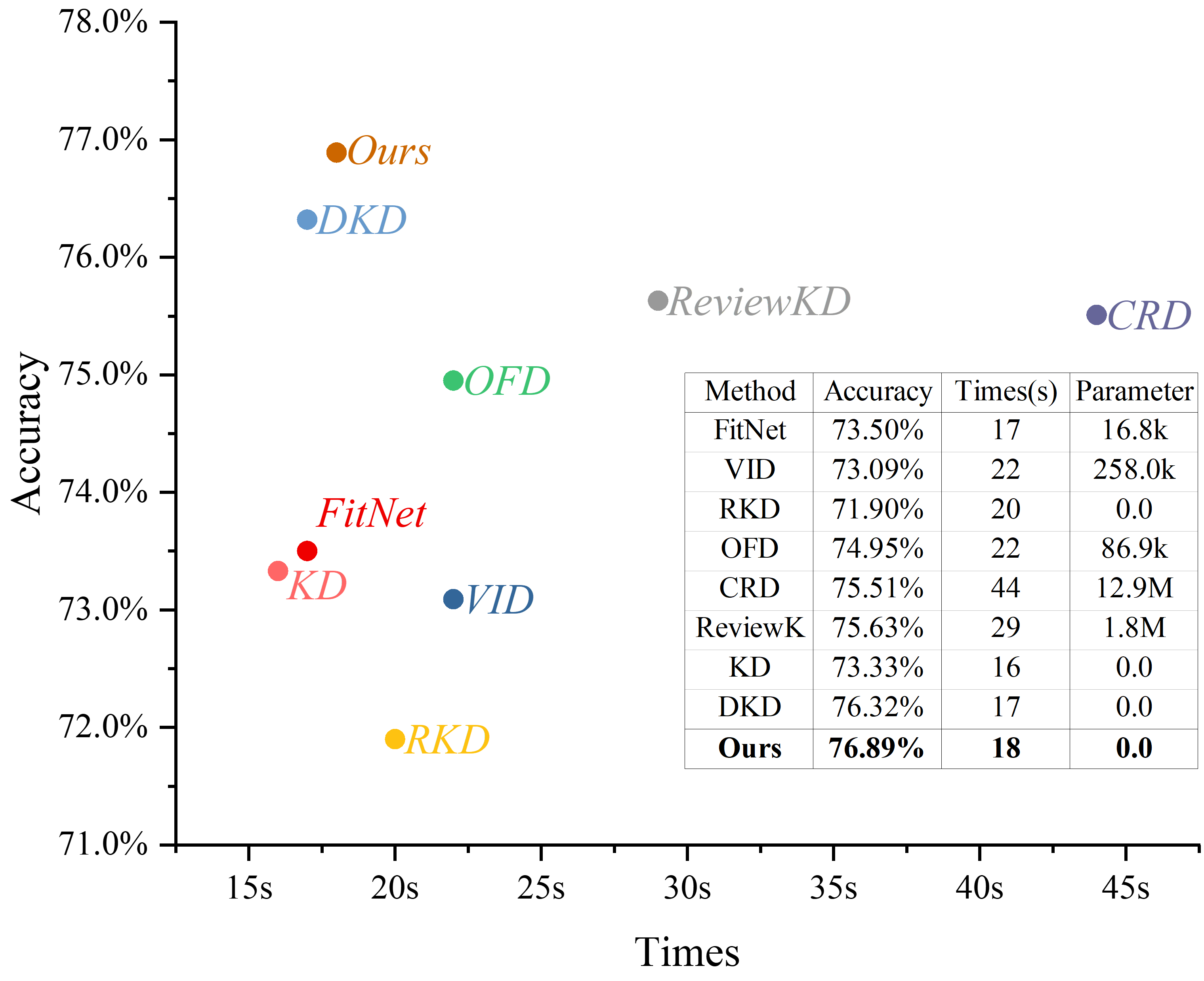} 
    \caption{Comparison of different KD methods in terms of training time, accuracy, and additional training parameters (on the CIFAR-100 dataset, using ResNet32x4-ResNet8x4 teacher-student model combination)}
	\label{fig:4-2-3} 
\end{figure}

\subsection{Robustness Experiments} To demonstrate that CMKD facilitates the student model's acquisition of clear and robust decision boundaries, as well as additional knowledge related to robustness and generalization, we conducted robustness tests on the student model using the CIFAR-100-C dataset \cite{hendrycks2019benchmarking}. The CIFAR-100-C dataset applies 15 different types of corruption, such as noise, blur, and occlusion, to the original CIFAR-100 images, with each type of corruption having five different severity levels to evaluate the model's robustness under these damaging conditions. Specifically, we assessed the robustness accuracy of the student model for five corruption types by calculating the average performance across the CIFAR-100-C images with five distinct corruption levels. The experimental results are shown in Table \ref{tab:4-3-1}. Compared to conventional KD, our method maintained a higher accuracy across the five different corruption types, demonstrating that our approach not only improves model performance but also enhances model robustness.

\begin{table*}
    \centering
    \caption{Robustness of KD and CMKD on the CIFAR-100-C dataset using five common corruption methods}
    \begin{tabular}{c|cccccc} 
    \hline
           & \multicolumn{6}{c}{Res32x4-Res8x4}                                                                        \\ 
    \hline
    Method & Clean   & brightness       & contrast          & elastic          & fog               & pixelate           \\ 
    \hline
    KD     & 73.33 & 64.88(-8.45) & 48.71(-24.62) & 55.3(-18.03) & 57.59(-15.74) & 45.26(-28.07)  \\
    ours   & 76.89 & 69.69(-7.20) & 55.52(-21.37) & 59.6(-17.29) & 63.81(-13.08) & 50.36(-26.53)  \\
    \hline
    \end{tabular}
    \label{tab:4-3-1}
    \end{table*}

\subsection{Ablation Studies}
In this section, we investigated the impact of the scaling coefficient as a hyperparameter on the overall performance, as well as the contribution of different components of our algorithm to the overall performance. All experiments were conducted on the CIFAR-100 dataset, and we selected two sets of teacher/student model combinations to verify the generalizability of the ablation study results. These combinations are ResNet32x4/ResNet8x4 and ResNet56/ResNet20.

\subsubsection{Hyperparameter} We set $\alpha=1$, $\gamma=1$ and vary the value of $\beta$ within the set \{1, 2, 3, 4, 5\} to identify the optimal hyperparameters. As shown in Figure \ref{fig:4-2-3}, when the $\beta$ is set to 4, both sets of teacher/student model combinations achieve significant performance improvements. Therefore, in this study, the hyperparameters are as follows: $\alpha=1$, $\beta=4$ and $\gamma=1$.

\begin{figure}[htbp] 
	\centering
	\includegraphics[width=\linewidth]{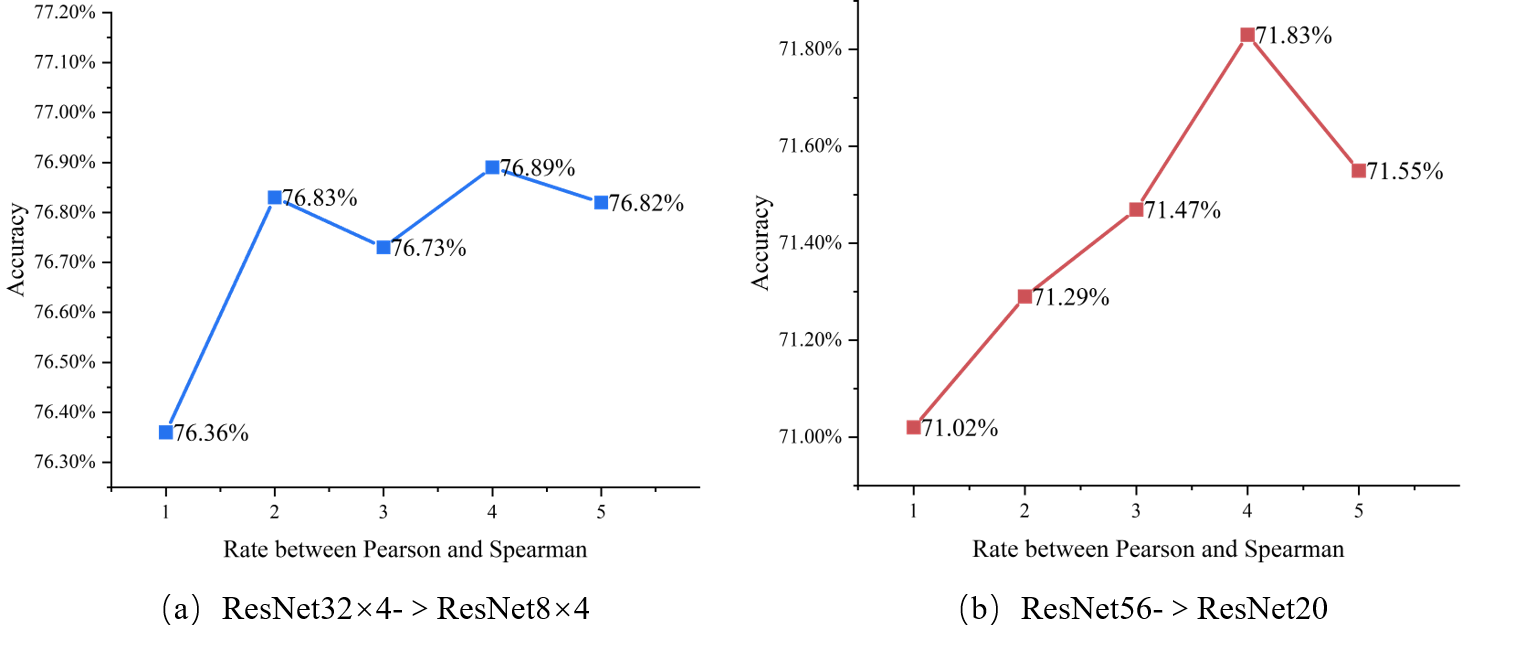} % The brackets adjust the size of the figure
    \caption{Impact of different scaling coefficients on model performance on the CIFAR-100 dataset}
	\label{fig:4-2-3} % Reference tag for citation in the document
\end{figure}

\subsubsection{Impact of Different Components} To demonstrate the effectiveness of each proposed component, we conducted ablation studies on the CIFAR-100 dataset. The results are shown in Table \ref{tab:4-4-2}. Compared to traditional KD, using the Pearson correlation coefficient instead of KL divergence alone resulted in improvements, with increases of 2.25\% and 0.39\% in two different teacher/student model combinations, respectively. When Pearson correlation coefficient is calculated with z-score normalization, the performance improvement was more significant, with an additional increase of 0.97\% and 0.44\%. Finally, by incorporating the Spearman correlation coefficient to capture additional knowledge information, the model performance was further enhanced, with further improvements of 0.34\% and 0.34\% compared to the previous methods.

\begin{table*}
    \centering
    \caption{Comparison of the impact of several key components in this paper on model performance on the CIFAR-100 dataset}
    %\resizebox{\columnwidth}{!}
    {%
    \begin{tabular}{cccc|c|c} 
    \hline
    \multicolumn{4}{c|}{Module}    & \multicolumn{2}{c}{Teacher-Student}       \\ 
    \hline
    KL & Pearson & Z-score Normalization & Spearman & ResNet32x4-ResNet8x4 & ResNet50-ResNet20  \\ 
    \hline
    $\times$& $\times$&$\times$ & $\times$ & 72.50                & 69.06\\
    \checkmark & $\times$ & $\times$ & $\times$ & 73.33                & 70.66              \\ 
    \hline
    $\times$ & \checkmark & $\times$ & $\times$ & 75.58(+2.25) & 71.05(+0.39) \\
    $\times$ & \checkmark & \checkmark & $\times$ & 76.55(+3.22) & 71.49(+0.83) \\
    $\times$ & \checkmark & \checkmark & \checkmark & 76.89(+3.56) & 71.83(+1.17) \\
    \hline
    \end{tabular} }
    \label{tab:4-4-2}
    \end{table*}

\begin{figure}[htbp] 
	\centering
	\includegraphics[width=\linewidth]{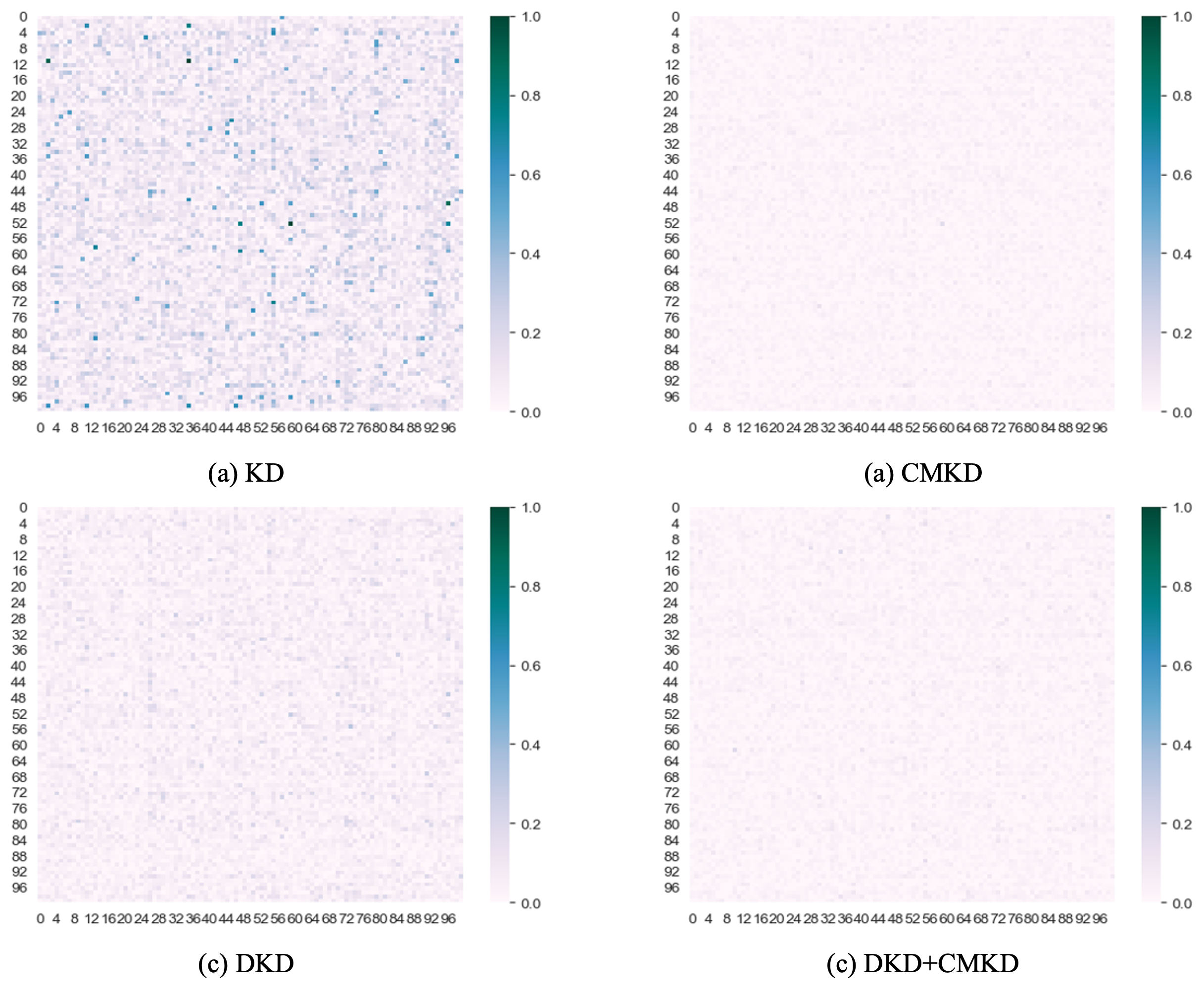} % The brackets adjust the size of the figure
    \caption{Correlation matrix of logits outputs for ResNet32x4/ResNet8x4 on the CIFAR-100 dataset}
	\label{fig:4-5-1} % Reference tag for citation in the document
\end{figure}

\begin{figure*}[htbp] 
	\centering
	\includegraphics[width=\linewidth]{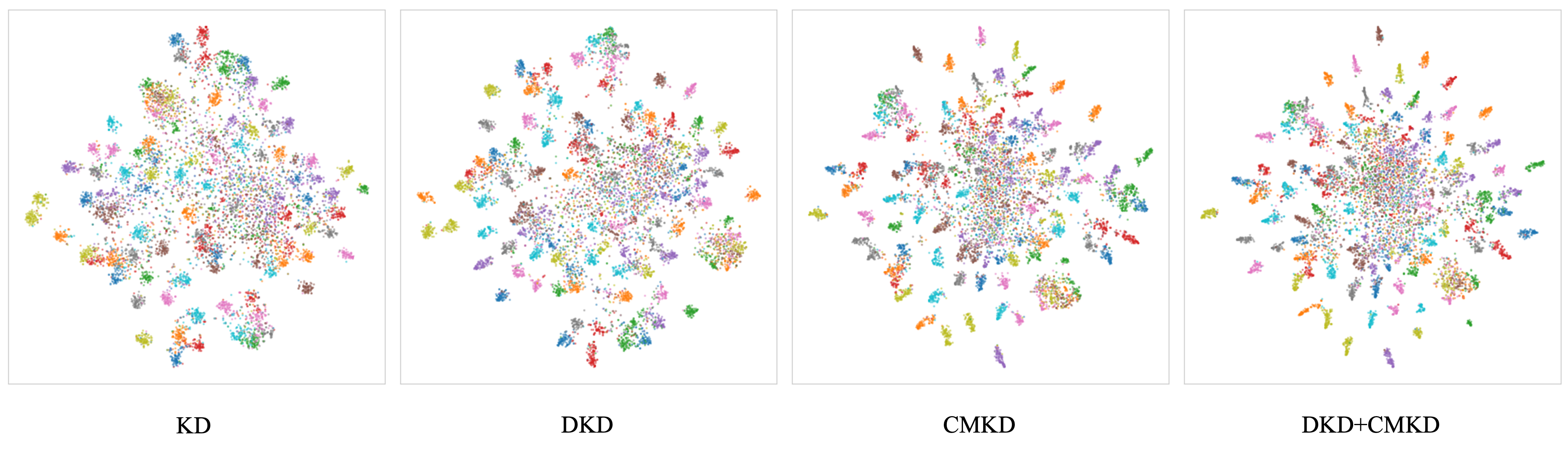} % The brackets adjust the size of the figure
    \caption{t-SNE dimensionality reduction visualization for ResNet32x4/ResNet8x4 on the CIFAR-100 dataset}
	\label{fig:4-5-2} % Reference tag for citation in the document
\end{figure*}

\subsection{Visualization}
To more intuitively demonstrate the effectiveness of our proposed method, we visualized the model using the ResNet32x4/ResNet8x4 as the teacher/student model on CIFAR-100 from two different perspectives. On the one hand, Figure \ref{fig:4-5-1} shows the difference in the correlation matrix of the global logits between the student and teacher. Darker colors indicate greater differences between the logits of the student and teacher. It can be observed that our method helps the student model to obtain more similar logit outputs from the teacher model, thus leading to better performance. On the other hand, we performed t-SNE visualizations, where different colors represent different categories in the classification. As shown in Figure~\ref{fig:4-5-2}, compared to traditional KD, CMKD demonstrates better separability, which proves that our proposed approach enhances the distinguishability and discriminative ability of the deep features in the student model.

\section{Conclusion}
\label{Section VII}
In this study, we provide a novel perspective on capacity mismatch through inter-class relationships. We empirically find that the KL-based KD method may implicitly change the inter-class relationships learned by the student model, resulting in a more complex and ambiguous decision boundary. To address this issue, we propose a novel correlation-based KD method (CMKD) that enables the student model to not only assimilate the value knowledge from the teacher's logits but also to emphasize the inter-rank knowledge inherent in the teacher's logits. Experimental results demonstrate that CMKD effectively reduces the discrepancy between the logits of the student and teacher models, facilitating the student's acquisition of robust knowledge from the more powerful and stronger teacher model.

\bibliography{IEEE-REF}{} 
\bibliographystyle{IEEEtran}

\appendix

In knowledge distillation, the loss function typically consists of two parts: one is the traditional cross-entropy loss, and the other is the KL divergence loss, which measures the difference between the output distributions of the teacher and student models. Here, we focus on the KL divergence component of the loss function, as shown in Eq. (\ref{eq17}).

\begin{equation}
    \label{eq17}
    \begin{aligned}
    \mathcal{L}_{KD} 
        & = \mathcal{L}_{KL}(\boldsymbol{p}^{s}(\tau), \boldsymbol{p}^{t}(\tau)) \\  
        & = \sum_{j} \boldsymbol{p}_{j}^{t}(\tau) \log \boldsymbol{p}_{j}^{t}(\tau) 
        - \sum_{j} \boldsymbol{p}_{j}^{t}(\tau) \log \boldsymbol{p}_{j}^{s}(\tau)
    \end{aligned}
\end{equation}
where, $\boldsymbol{p}_{j}^{t}(\tau) = \frac{e^{\boldsymbol{z}_{j}^{t}/\tau}}{\sum_{i} e^{\boldsymbol{z}_{i}^{t}/\tau}}$ is the softened output probability of the teacher model for class \( i \) (with temperature \( \tau \)). Similarly, $\boldsymbol{p}_{j}^{s}(\tau) = \frac{e^{\boldsymbol{z}_{j}^{s}/\tau}}{\sum_{i} e^{\boldsymbol{z}_{i}^{s}/\tau}}$ is the softened output probability of the student model for class \( i \) (with temperature \( \tau \)).
\( z_i \) and \( s_i \) are the logits of the teacher and student models for class \( i \), respectively.

Next, we derive the gradient of the KL divergence loss $\mathcal{L}_{KD}$ with respect to the logits \( z_k^s \) of the student model for the input of class \( k \), following the chain rule.

\begin{equation}
    \label{eq18}
    \begin{aligned}
    \frac{\partial \mathcal{L}_{KD}}{\partial z_k^s} 
        & = \sum_{j} -p_{j}^{t}(\tau) \frac{\partial \log p_j^s(\tau)}{\partial z_k^s} \\  
        & = \sum_{j} -\frac{p_j^t(\tau)}{p_j^s(\tau)} \frac{\partial p_j^s(\tau)}{\partial z_k^s} \\
        & = -\frac{\boldsymbol{p}_k^t(\tau)}{\boldsymbol{p}_k^s(\tau)} \frac{\partial \boldsymbol{p}_k^s(\tau)}{\partial \boldsymbol{z}_k^s}
        - \sum_{j \neq k} \frac{\boldsymbol{p}_j^t(\tau)}{\boldsymbol{p}_k^s(\tau)} \frac{\partial \boldsymbol{p}_j^s(\tau)}{\partial \boldsymbol{z}_k^s}
    \end{aligned}
\end{equation}

For $\frac{\partial \boldsymbol{p}_j^s(\tau)}{\partial \boldsymbol{z}_k^s}$ in Eq. (\ref{eq18}),
substituting $\boldsymbol{p}_{j}^{s}(\tau) = \frac{e^{\boldsymbol{z}_{j}^{s}/\tau}}{\sum_{i} e^{\boldsymbol{z}_{i}^{s}/\tau}}$, we get:
\begin{equation}
    \label{eq19}
    \begin{aligned}
    \frac{\partial p_{j}^{s}(\tau)}{\partial z_{k}^{s}} = & \frac{\partial}{\partial z_{k}^{s}} \left(\frac{e^{z_{j}^{s}/\tau}}{\sum_{i} e^{z_{i}^{s}/\tau}}\right) \\  
        & = \frac{\frac{\partial}{\partial z_{k}^{s}} \left(e^{z_{j}^{s}/\tau}\right) \sum_{i} e^{z_{i}^{s}/\tau} - e^{z_{j}^{s}/\tau} \frac{\partial}{\partial z_{k}^{s}} \left(\sum_{i} e^{z_{i}^{s}/\tau}\right)}{\left(\sum_{i} e^{z_{i}^{s}/\tau}\right)^2}
    \end{aligned}
\end{equation}

When \( j = k \), only the terms related to \( j \) and \( k \) are non-zero. Applying the chain rule, the result of Eq. (\ref{eq19}) is:
\begin{equation}
    \label{eq20}
    \begin{aligned}
    \frac{\partial p_{j}^{s}(\tau)}{\partial z_{k}^{s}} 
        & = \frac{\partial}{\partial z_{k}^{s}} \left(\frac{e^{z_{j}^{s}/\tau}}{\sum_{i} e^{z_{i}^{s}/\tau}}\right) \\  
        & = \frac{\frac{1}{\tau} e^{z_{j}^{s}/\tau} \sum_{i} e^{z_{i}^{s}/\tau} - e^{z_{j}^{s}/\tau} \frac{1}{\tau} e^{z_{j}^{s}/\tau}}{\left(\sum_{i} e^{z_{i}^{s}/\tau}\right)^2} \\
        & = \frac{1}{\tau} \frac{e^{z_{j}^{s}/\tau}}{\sum_{i} e^{z_{i}^{s}/\tau}} \left(1 - \frac{e^{z_{j}^{s}/\tau}}{\sum_{i} e^{z_{i}^{s}/\tau}}\right)
    \end{aligned}
\end{equation}

Since $\boldsymbol{p}_{j}^{s}(\tau) = \frac{e^{\boldsymbol{z}_{j}^{s}/\tau}}{\sum_{i} e^{\boldsymbol{z}_{i}^{s}/\tau}}$, Eq. (\ref{eq20}) simplifies to:
\begin{equation}
    \label{eq21}
    \frac{\partial p_{j}^{s}(\tau)}{\partial z_{k}^{s}} = \frac{1}{\tau} \boldsymbol{p}_k^s(\tau) \left(1 - \boldsymbol{p}_k^s(\tau)\right)
\end{equation}

Similarly, for \( j \neq k \), applying the chain rule, the result of Eq. (\ref{eq19}) is:

\begin{equation}
    \label{eq22}
    \begin{aligned}
    \frac{\partial p_j^s(\tau)}{\partial z_k^s} 
        & = \frac{\partial}{\partial z_k^s} \left(\frac{e^{z_j^s/\tau}}{\sum_{i} e^{z_i^s/\tau}}\right) \\  
        & = -\frac{e^{z_{j}^{s}/\tau} \frac{1}{\tau} e^{z_{k}^{s}/\tau}}{\left(\sum_{i} e^{z_{i}^{s}/\tau}\right)^2} \\
        & = -\frac{1}{\tau} \frac{e^{z_j^s/\tau}}{\sum_{i} e^{z_i^s/\tau}} \frac{e^{z_k^s/\tau}}{\sum_{i} e^{z_i^s/\tau}}
    \end{aligned}
\end{equation}

Again, substituting $\boldsymbol{p}_{j}^{s}(\tau) = \frac{e^{\boldsymbol{z}_{j}^{s}/\tau}}{\sum_{i} e^{\boldsymbol{z}_{i}^{s}/\tau}}$ and $\boldsymbol{p}_{k}^{s}(\tau) = \frac{e^{\boldsymbol{z}_{k}^{s}/\tau}}{\sum_{i} e^{\boldsymbol{z}_{i}^{s}/\tau}}$, Eq. (\ref{eq22}) simplifies to:

\begin{equation}
    \label{eq23}
    \frac{\partial p_j^s(\tau)}{\partial z_k^s} = \frac{1}{\tau} p_{j}^{s}(\tau) p_{k}^{s}(\tau)
\end{equation}

Finally, substituting Eqs. (\ref{eq21}) and (\ref{eq23}) into Eq. (\ref{eq18}), and simplifying the expression, we obtain:

\begin{equation}
    \label{eq24}
    \frac{\partial \mathcal{L}_{KD}}{\partial z_k^s}  = \frac{1}{\tau} \left[p_{k}^{s}(\tau) - p_{k}^{t}(\tau)\right]
\end{equation}

\vspace{11pt}
%\bf{Authors:}\vspace{-33pt}
%\begin{IEEEbiography}
%[{\includegraphics[width=1in,height=1.25in,clip,keepaspectratio]{fig/wang.png}}]{Yingchao Wang} is currently pursuing a Ph.D. degree with the School of Cyberspace Science and Technology, Beijing Institute of Technology, Beijing, China.

%His research interests include machine learning, deep learning, knowledge distillation, and end-edge-cloud computing. 

%\end{IEEEbiography}

%\begin{IEEEbiography}
%[{\includegraphics[width=1in,height=1.25in,clip,keepaspectratio]{fig/niu.png}}]{Wenqi Niu} has received the M.S. degree in Electronic Information Science and Technology from North Minzu Universit, NingXia, Yinchuan, China, in 2021. 

%His research interests include deep learning, knowledge distillation, image processing and detection.

%\end{IEEEbiography}

\vfill

\end{document}